\documentclass[lettersize,journal,onecolumn]{IEEEtran}
\usepackage{mathptmx} % assumes new font selection scheme installed
\usepackage{times} % assumes new font selection scheme installed
\usepackage[dvipdfmx]{graphicx}
\usepackage{url}
\usepackage{graphicx}
\usepackage[figurename=Fig.\ ]{caption}
\usepackage{epsfig} % for postscript graphics files
\usepackage{cite}
\usepackage{color}
\usepackage{booktabs}  % 
\usepackage{amsmath}
\usepackage{amssymb}
\usepackage{indentfirst}
\usepackage{algorithm}
\usepackage{algorithmicx}
\usepackage{algpseudocode}
\usepackage{caption}
\usepackage{subcaption}
\usepackage{comment}
\usepackage{color,soul}
\usepackage{ifthen}
\usepackage[table,xcdraw]{xcolor}
\usepackage{hyperref}
\usepackage{animate}
\title{\LARGE \bf
GPT Models Meet Robotic Applications: Co-Speech Gesturing Chat System
}

\author{
Naoki Wake$^{1}$,
Atsushi Kanehira$^{1}$,
Kazuhiro Sasabuchi$^{1}$,
Jun Takamatsu$^{1}$,
and Katsushi Ikeuchi$^{1}$
\thanks{$^{1}$Applied Robotics Research, Microsoft, 
        Redmond, WA 98052, USA
        {\tt\small naoki.wake@microsoft.com}}%
}

\begin{document}
\maketitle
\thispagestyle{empty}
\pagestyle{empty}

\begin{abstract}
This technical paper introduces a chatting robot system that utilizes recent advancements in large-scale language models (LLMs) such as GPT-3 and ChatGPT (Fig.\ref{fig:eyecatch}). The system is integrated with a co-speech gesture generation system, which selects appropriate gestures based on the conceptual meaning of speech. Our motivation is to explore ways of utilizing the recent progress in LLMs for practical robotic applications, which benefits the development of both chatbots and LLMs. Specifically, it enables the development of highly responsive chatbot systems by leveraging LLMs and adds visual effects to the user interface of LLMs as an additional value. The source code for the system is available on \href{https://github.com/microsoft/LabanotationSuite/tree/master/MSRAbotChatSimulation}{GitHub for our in-house robot} and \href{https://github.com/microsoft/GPT-Enabled-HSR-CoSpeechGestures}{GitHub for Toyota HSR}.
\end{abstract}

\section{Introduction}
In recent years, large-scale language models (LLMs) such as GPT-3~\cite{brown2020language} and ChatGPT~\cite{OpenAI} have achieved remarkable success in natural language processing tasks. Meanwhile, there is a growing interest in applying LLMs to robotic applications (task planning, for example,~\cite{wake2023chatgpt}), as they have the potential to enhance human-robot interaction with its high language processing capabilities.

In this paper, we present a co-speech gesturing chat system that combines the GPT-3 model with a gesture engine developed by our team\cite{wake2022rinna, teshima2022integration}. The gesture engine analyzes the text of the robot's speech and selects an appropriate gesture from a library of gestures associated with the conceptual meaning of the speech, called as gesture library\cite{ikeuchi2019design}. By integrating LLMs as the backend, we aim to provide users with a highly responsive chat system that can handle a wider range of conversational topics and contexts. %Additionally, we developed the pipeline utilizing various Azure services\cite{azure-services} to enhance the functionality and efficiency of our system. Specifically, we utilized Azure speech service for speech-to-text conversion, Azure Open AI's GPT-3 model for generating responses, and Azure language understanding for concept estimation. 

We believe that LLMs will contribute significantly to the development of practical robotic systems. Taking chatting robots as an example, developers can develop highly responsive systems with simple prompt engineering. Another interesting topic will be how adding visual effects to the user interface of LLMs, which are mostly text-based interactions, will affect the usability and conversational content. We have implemented the chat system into two robots, our in-house MSRAbot and Toyota HSR. The source code for the system is available on \href{https://github.com/microsoft/LabanotationSuite/tree/master/MSRAbotChatSimulation}{GitHub for our in-house robot} and \href{https://github.com/microsoft/GPT-Enabled-HSR-CoSpeechGestures}{GitHub for Toyota HSR}.

\section{Pipeline}
\begin{figure*}[tb]
  \centering
  \includegraphics[width=0.8\textwidth]{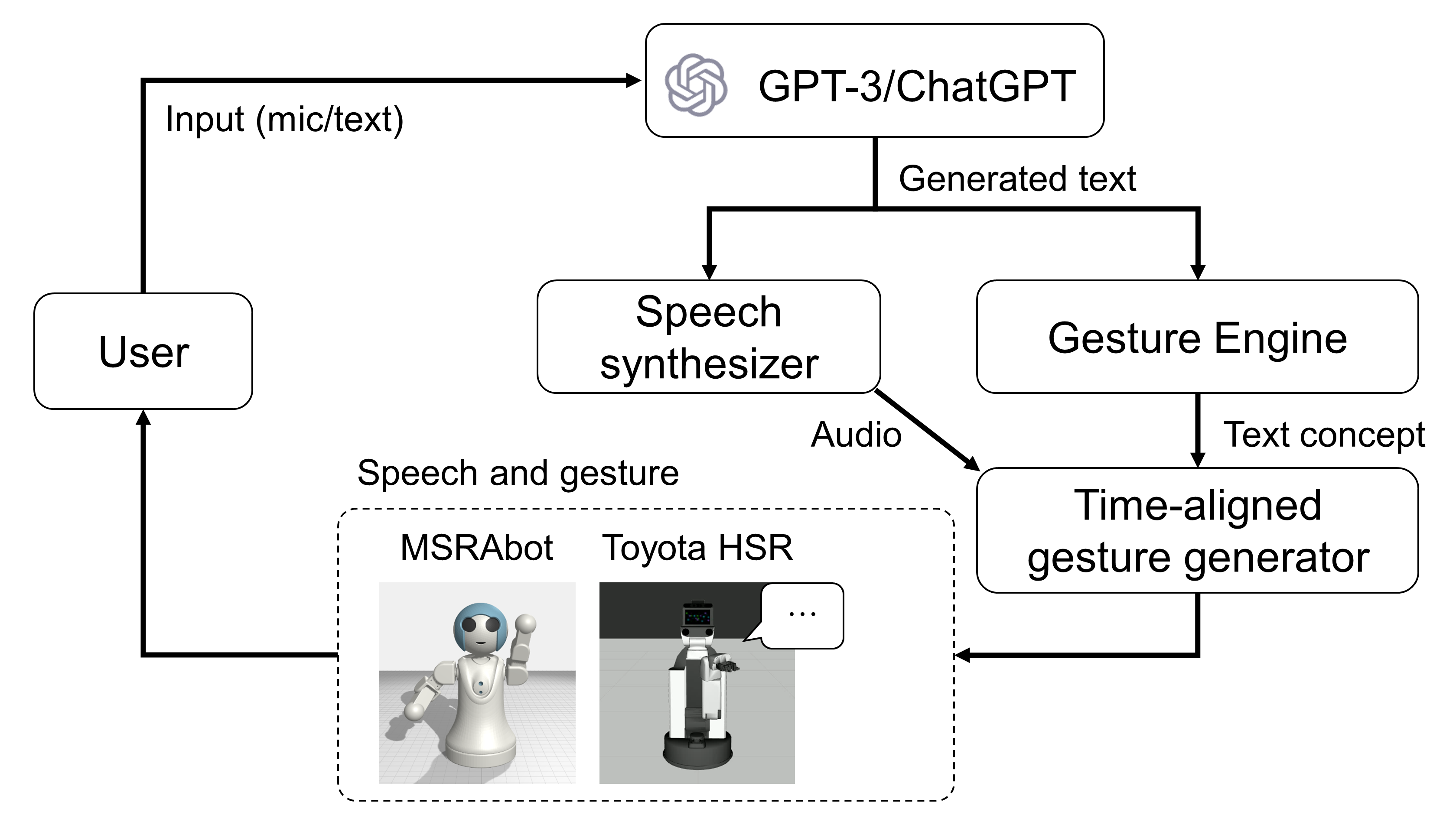}
  \caption{
  The pipeline of the co-speech gesture generation system.
  }
  \label{fig:title}
  % \vspace{-15pt}
\end{figure*}

The overview of the pipeline is shown in Fig.\ref{fig:title}. 
The user sends a query to the robot system via text or microphone input. Microphone input is noise-suppressed to prevent the robot's ego noise from interfering with recognition \cite{wake2019enhancing,jaroslavceva2022robot}and then converted to text using a third-party text-to-speech technology~\cite{azure-services}. The robot system then generates a prompt for the GPT-3/ChatGPT model based on this input.

\subsection{Chatting engine}
\subsubsection{GPT-3 model as backend}
As GPT-3 is not specialized for chat and is designed for text completion, the prompt needs to be carefully crafted to achieve chat-like responses. To do this, we save the conversation history between the user and the robot system and design the following prompt:

``You are an excellent chat bot. Please respond to the current message accurately, taking into account your knowledge and our previous conversations.
Previous conversations: \textit{history}
Current message: \textit{message}''

\begin{figure*}[tb]
  \centering
  \includegraphics[width=0.8\textwidth]{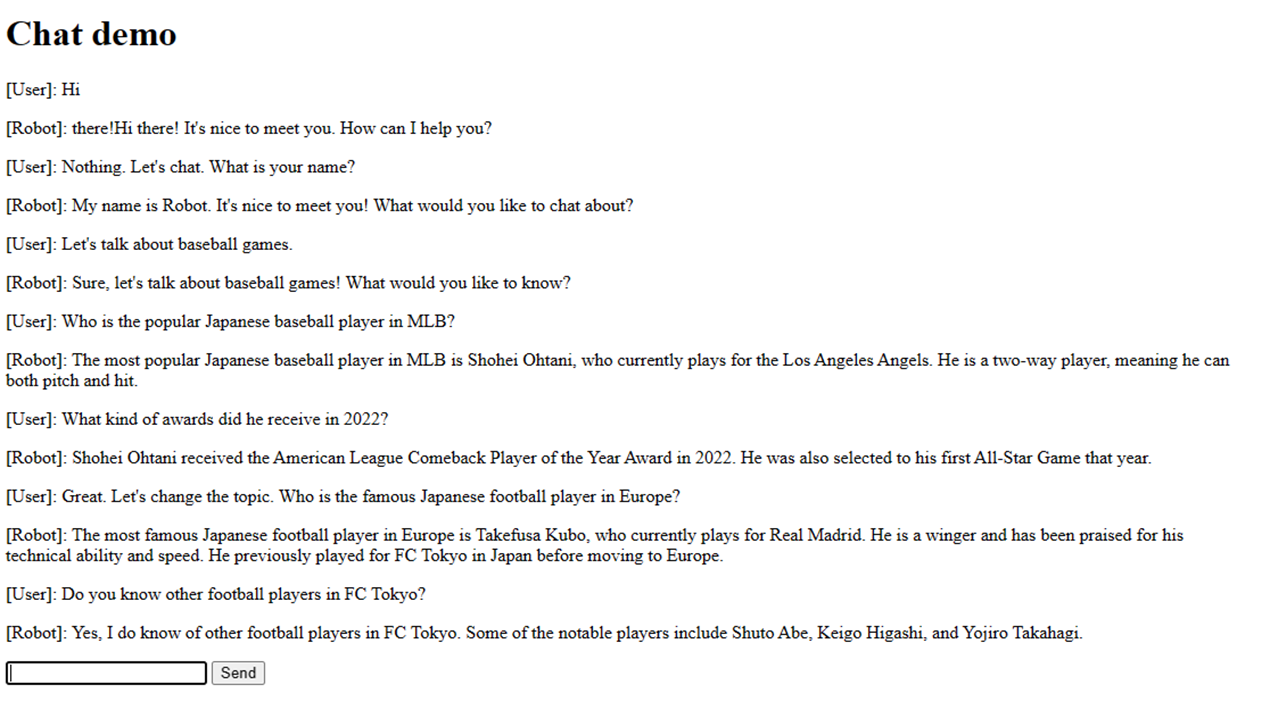}
  \caption{
  Example of the conversation history between a human user and the chat system.
  }
  \label{fig:gptexample}
  % \vspace{-15pt}
\end{figure*}

Here, \textit{history} is assigned the conversation exchange separated by line breaks, and \textit{message} is assigned the new input from the user. We used the Azure Open AI's davinci model as the specific model. The example of the conversation is shown in Fig.\ref{fig:gptexample}.

\subsubsection{ChatGPT model as backend}
We used the Azure Open AI's gpt-3.5-turbo model as the specific model. Since the ChatGPT model can receive conversation history, we simply described roles in the prompts, without embedding the conversation history. The following is an example of a prompt we prepared:

``You are an excellent chat bot, named MSRAbot. You are embodied with a small robot, which makes lively gestures in response to your speech. Please keep conversations with the user by responding with short English phrases. The response can be composed of several sentences, but every sentence should be definitely short and less than 12 words. Answer in English in any situation.''

\subsection{Gesture engine}
The response from GPT-3/ChatGPT is passed on to two modules: a speech generator, which converts the text into speech using a third-party text-to-speech technology~\cite{azure-services}, while the gesture engine selects a concept from the text. For concept estimation, we preliminarily analyzed a collection of everyday English conversation phrases and defined dozens of concepts that are commonly represented in conversations~\cite{ikeuchi2019design}. Based on the concept labels for the phrases, we trained the estimation model using Azure Language Understanding~\cite{azure-services}.

The gesture generator then produces gestures based on the estimated concept. When several gestures are associated with the concept, we randomly select one. The length of the gesture was modified to match the length of synthesized speech. Finally, the generated speech and co-speech gestures are presented to the user as audio-visual feedback.

\subsection{Open sources}
Note that we store the information of gesture motions using Labanotation, a notation for describing human dance. Labanotation is an intermediate representation of human motions that contains enough information to reproduce them. Thus, by implementing a Labanotation decoder for each robot, this pipeline can be scaled to arbitrary robots~\cite{ikeuchi2018describing}. Currently, we have prepared the decoder for our in-house robot, MARabot, and Toyota HSR, and we have open-sourced the code on \href{https://github.com/microsoft/LabanotationSuite/tree/master/MSRAbotChatSimulation}{GitHub for our in-house robot} and \href{https://github.com/microsoft/GPT-Enabled-HSR-CoSpeechGestures}{GitHub for Toyota HSR}. MSRAbot was originally designed as a platform for human-robot interaction research, and we have developed and open-sourced a \href{https://github.com/microsoft/gestureBotDesignKit}{DIYKit}. This DIYKit includes 3D models of the parts and step-by-step assembly instructions, enabling users to build the robot's hardware using commercially available items. The software needed to operate the robot is also available on the same page.

\begin{figure*}[tb]
  \centering
  \includegraphics[width=0.8\textwidth]{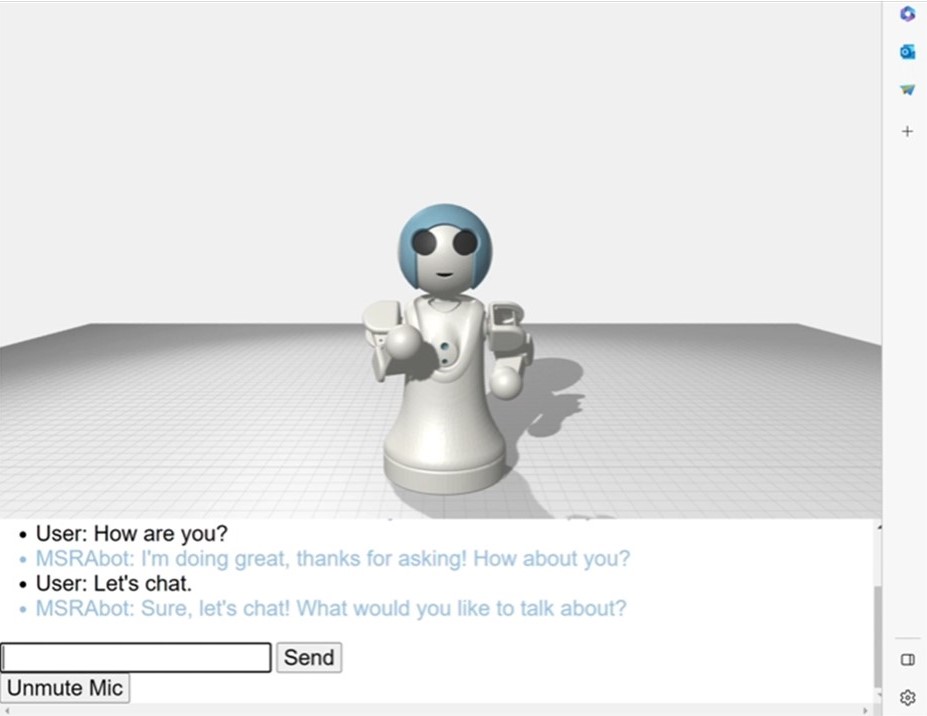}
  \caption{Our robotic gesture engine and DIY robot, MSRAbot, are integrated with a GPT-based chat system.}
  \label{fig:eyecatch}
\end{figure*}

\section{Discussions and conclusion}
LLMs like GPT-3/ChatGPT have shown remarkable success in natural language processing tasks, leading to growing interest in applying them to robotic applications. However, connecting a robot with LLMs poses risks such as bias, inappropriate responses, or vulnerability to attacks. Solutions to these problems are in the process of development. To minimize those risks, it is crucial to carefully monitor and control the robot's output, utilize robust security measures, and provide proper ethical guidelines. %On the other hand, utilizing cloud services like Azure offers several benefits, such as efficient and scalable processing power, advanced AI models, and reliable data storage.

In conclusion, this paper introduced an LLM-empowered chatting robot system for achieving a natural and intuitive chatting experience, while also providing synchronized co-speech gestures. We believe that LLMs will facilitate the development of practical robotic applications, provided that we pay close attention to the limitations of those models.
\bibliographystyle{ieeetr}
\bibliography{bib}

\end{document}